# Examining Scientific Writing Styles from the Perspective of Linguistic Complexity

Chao Lu[1, 2], Yi Bu[3], Jie Wang[4], Ying Ding[2,5], Vetle Torvik[6], Matthew Schnaars[2], Chengzhi Zhang[1,7], *

1. School of Economics and Management, Nanjing University of Science and Technology, Nanjing, Jiangsu, China

2. School of Informatics, Computing, and Engineering, Indiana University, Bloomington, Indiana, U.S.A.

3. Center for Complex Networks and Systems Research, School of Informatics, Computing, and Engineering, Indiana University, Bloomington, Indiana, U.S.A.

4. School of Information Management, Nanjing University, Nanjing, Jiangsu, China

5. School of Information Management, Wuhan University, Wuhan, Hubei, China

6. School of Information Sciences, University of Illinois, Urbana, Illinois, U.S.A.

7. Jiangsu Key Laboratory of Data Engineering and Knowledge Service, Nanjing University, Nanjing, Jiangsu, China

**Corresponding author: Chengzhi Zhang** (zhangcz@njust.edu.cn).

# Examining Scientific Writing Styles from the Perspective of Linguistic Complexity

**Abstract:** Publishing articles in high-impact English journals is difficult for scholars around the world, especially for non-native English-speaking scholars (NNESs), most of whom struggle with proficiency in English. In order to uncover the differences in English scientific writing between native English-speaking scholars (NESs) and NNESs, we collected a large-scale data set containing more than 150,000 full-text articles published in *PLoS* between 2006 and 2015. We divided these articles into three groups according to the ethnic backgrounds of the first and corresponding authors, obtained by Ethnea, and examined the scientific writing styles in English from a two-fold perspective of linguistic complexity: (1) syntactic complexity, including measurements of sentence length and sentence complexity; and (2) lexical complexity, including measurements of lexical diversity, lexical density, and lexical sophistication. The observations suggest marginal differences between groups in syntactical and lexical complexity.

## 1. INTRODUCTION

*Background*

When we discuss publishing papers in a high quality journal, English-language journals, like *Nature*, *Science*, *PNAS,* are usually the first examples that come to mind. According to the Journal Citation Report (JCR) released in 2016, 8,778 journals were indexed in SCI (Science Citation Index) or SSCI (Social Science Citation Index); those published in the United States and the United Kingdom made up more than half (33.8%+20.5%, respectively), not even including English journals published in non-English-speaking countries, as shown in Table 1. Since the number of publications in journals indexed by SCI/SSCI is important in evaluating the scientific outputs of scholars, publishing papers in English becomes a major criterion to measure individual scholars' academic achievements.

**Table 1. Geographical distribution of the journals in JCR 2016 (Top 10).**

| Country | Count | Ratio (%) |
|---------|-------|-----------|
| UNITED STATES | 2,966 | 34 |
| UNITED KINGDOM | 1,796 | 20 |
| NETHERLANDS | 712 | 8 |
| GERMANY | 581 | 7 |
| JAPAN | 225 | 3 |
| SWITZERLAND | 208 | 2 |
| FRANCE | 181 | 2 |
| PEOPLES R CHINA | 181 | 2 |
| RUSSIA | 136 | 2 |
| POLAND | 134 | 2 |

Non-native English-speaking scholars (NNESs) inevitably face more challenges when publishing articles in English compared with native English-speaking scholars (NESs) because of the language barrier. Therefore, numerous researchers have studied possible problems hindering NNESs from publishing in English journals (e.g., Ferris, 1994a, 1994b; Flowerdew, 1999). Of these studies, language proficiency is the most discussed. For example, all the interviewed non-native English Ph.D. students in Taiwan acknowledged that language barrier prevents them from publishing manuscripts in English journals (Huang, 2010). Additionally, the editors of a medical journal commented on the weaknesses in scientific English writing of NNESs (Mišak, Marušić, & Marušić, 2005). These scholars thus sometimes fail to meet the expectation of the reviewers by great margins (Curry & Lillis, 2004). Furthermore, the reviewers' potential subconscious biases against both papers demonstrating poor

English proficiency and NNESs as a group may exacerbate the already difficult situation of NNESs (Flowerdew, 2000; Tomkins, Zhang, & Heavlin, 2017). Even when their articles are accepted for publication, the reviewers could still request further language improvement (Huang, 2010). In these cases, NNESs usually refer to professional help—a paid editing service—to increase the likelihood of publication (Bailey, 2011), especially for those with no opportunity to collaborate with NESs. These paid services are usually expensive and provide final drafts without detailed explanation. Even when they provide editorial feedback, the resultant improvements of language proficiency for the NNES author(s) are limited (Chandler, 2003; Ferris, 2004).

Nowadays, collaboration across groups, labs, and disciplines is becoming almost inevitable (Zhang *et al.*, 2018). For example, among 155,579 articles published in *PLoS* journals, 108,531 (69.8%) are multi-authored publications. Numerous studies have already proved that scientific collaboration can improve the impact of their scientific publications (e.g., Lee & Bozeman, 2005). Therefore, analyzing collaborative writing is critical, especially when coauthors are either NNESs, NESs, or a mix of both (Dobao, 2012; Gebhardt, 1980). For example, in collaborative writing between advisors and advisees, studies suggest that advisors have great impact on the growth of advisees, including topic selection and advisees' writing skills (Huang, 2010; Shaw, 1991). Kessler, Bikowski, & Boggs (2012) found that NNESs wrote more accurately with collaboration. During collaborative writing, collaborators can gain better scientific knowledge, more skilled scientific reasoning, and improved writing (Mason, 1998; Jang, 2007).

*Objective*

In this article, we use data-driven approaches with a large-scale full-text data set of scientific articles to understand scientific English writing from the perspective of linguistic complexity with special focus on the collaborative writing of authors from different ethnic backgrounds. Based on the literature, we identify three indicators established in Second Language Acquisition of Linguistic Studies: Complexity, Accuracy, and Fluency (CAF). These three indicators have been widely adopted to assess the English proficiency of non-native English-speaking writers (NNEWs) or to compare differences between native English-speaking writers (NEWs) and NNEWs, especially in reading and writing (Beers & Nagy, 2009; Ellis & Yuan, 2004; Skehan, 2009). In general, Accuracy is usually measured by the number or ratio of errors (e.g. grammatical errors or lexical errors) to word count from the text or speech of an NNEW (Chandler, 2003; Ojima, 2006). Fluency is usually measured across time, such as syllables per minute or text length over a period of time (Skehan, 2009). But both indicators fail to capture English scientific writing styles for two reasons: accuracy fails because articles should be edited for errors before publication; and fluency does not apply because we cannot accurately obtain the authors' time spent on articles. By contrast, complexity can be promising for highlighting the differences between NESs and NNESs in English scientific writing. It has been used to measure NEWs' English Proficiency on its own (Ellis & Yuan, 2004; Ferris, 1994a, 1994b) for its advantage that it can be measured from the text alone. Additionally, the framework of linguistic complexity is relatively comprehensive, as it includes syntactic and lexical complexity with various quantitative variables (detailed in Methodology), well suited for large-scale data sets. Therefore, although we have identified three indicators, only one of them is applied to the empirical study.

This paper is organized as follows: section two comprises a brief literature review; section three presents our data set and the method for measuring the writing style with linguistic complexity; section four describes the results of the empirical study; section five discusses the results; and, lastly, section six draws conclusions and suggests future research.

## 2. RELATED WORK

*NNESs and Scientific Writing in English*

NNESs face various language problems in general English writing (Ferris, 1994a). Their writings have been concluded to be "less fluent (fewer words), less accurate (more errors), and less effective (lower holistic scores)" compared with NESs (Silva, 1993, p.668). For example, Ortega (2003) investigated the syntactic complexity of NNEWs' writing in a synthesis of 27 formal studies on college-level NNEWs and found that writers with better language proficiency would write significantly longer sentences. Likewise, Ferris (1994a) compared the persuasive writing by NEWs and NNEWs with different levels of language proficiency. Findings based on the analysis of 33 quantitative, rhetorical, and topical-structure variables indicated significant differences; for instance, NEWs tended to write longer sentences and compositions than NNEWs. Rabinovich, Nisioi, Ordan, and Wintnerb (2016) found that NEWs employ more pronouns and collocations than NNEWs in their articles. Other problems in NNEWs may also include inadequate content, poor organization, and stylistic inappropriateness (Crowhurst, 1991).

Furthermore, they also bore the additional weight of scientific writing (e.g., Flowerdew, 1999; Huang, 2010). After interviewing 26 participants in Hong Kong (most of whom were assistant professors), Flowerdew (1999) found that NNESs not only had the aforementioned problems (e.g. simple styles, less rich vocabularies, and side-effects from their first-language culture) but also had difficulty in writing the Introduction and Discussion that both require authors' critical thinking. By interviewing Ph.D. students in Taiwan, Huang (2010) found that although acknowledging their disadvantages in English scientific writing in their research, they had limited motivations to improve their English due to the imbalanced relationship between advisors and advisees in which advisees often found it difficult to assume full control of and responsibility in their work.

*Scientific Collaborative Writing*

Studies have shown that collaborative writing, compared with individual writing, produces written text that is more complex, accurate, and fluent (i.e., CAF), for both NEWs and NNEWs (Dobao, 2012; Gebhardt, 1980). Yarrow & Topping (2001) found among teenagers that collaborative writing created significantly better texts than individual writing and that within groups, collaborators with better writing skills tutored those with poorer skills during interactions, which indicates the benefits of collaborative writings between authors with different levels of language proficiencies. More specifically, in scientific collaborative writing, similar cases have been observed (Jang, 2007). Kessler, Bikowski, & Boggs (2012), for example, found that NNESs wrote more accurately with collaboration. Other studies have also shown that during discussion and writing, students developed better scientific knowledge, more skilled scientific reasoning, and improved writing (Keys, 1994; Mason, 1998; Jang, 2007), all of which are critical to the long-term improvement of scientific writing. However, despite the obvious importance of collaboration, these small-sampled findings can hardly paint the comprehensive view, because other factors need to be considered, e.g., linguistic disparities between collaborators, costs of coordination, and types of collaborations (Lowry, Curtis, & Lowry, 2004). Thus, further investigation is necessary to better understand collaborative writing in scientific collaboration.

*Quantitative Measurement of English Writing*

While sociolinguists use qualitative methods like interviews, experts in computational linguistics tend to adopt quantitative methods, such as natural language processing (NLP) technologies, in their studies (e.g., Brants, 2000; Brown *et al.*, 1993). The CAF indicators have been widely adopted in measuring English proficiency, especially with regards to writing (Ellis & Yuan, 2004). As addressed in previous sections, Complexity has various advantages over the others in measuring English scientific writing style. Complexity comprises two aspects: syntactic and lexical complexity.

*Syntactic complexity* consists of quantitative variables on sentence length, sentence complexity, and others (Ferris, 1994a; Kormos, 2011; Ojima, 2006). Lu (2010), for instance, selected 14 syntactic complexity measurements (including length of production unit, sentence complexity, subordination, coordination, and particular structures, etc.) and constructed a computational system for automatic analysis of syntactic complexity in second language writing. Campbell and Johnson (2001) compared the syntactic complexity in medical and non-medical corpora and argued that syntax of medical language shows less variation than the non-medical language. Recent studies concerning syntactic complexity have mainly demonstrated differences between specific language systems (i.e. linguistic families) (Yang, Marslenwilson, & Bozic, 2017; Scontras, Badecker, & Fedorenko, 2017). *Lexical complexity* is made up of lexical diversity, lexical density, and lexical sophistication (Vajjala & Meurers, 2012). Each of these variables have been used to measure writing from NNEWs or to compare the differences between NEWs and NNEWs (Ferris, 1994a; Ortega, 2003) These features have also been adopted in authorship attribution identification (Holmes, 1994), readability classification (Vajjala & Meurers, 2012), and gender identification in scientific articles (Bergsma, Post, & Yarowsky, 2012). This study applies these variables to describe the writing style of English scientific articles from different linguistic backgrounds.

*NLP-based Native Language Identification*

Native Language Identification (NLI), a task aiming to identify a person's first or native language (Jarvis & Paquot, 2015; Nisioi, 2015; Tsvetkov et al., 2013), is essentially a classification problem. NLI has been widely applied to speech recognition, parsing and information extraction (Mayfield & Jones, 2001), and author identification and profiling (Wong & Dras, 2011). Current NLP-based NLI studies heavily rely on the quality and coverage of the corpus (Jarvis & Paquot, 2015). Koppel et al. (2005), for example, used part-of-speech (POS) bigrams, character n-grams, and feature function words; their empirical study focused on five groups of NNESs and got an 80% accuracy rate in the NLI task. Estival et al. (2007) utilized lexical and structural features and raised the accuracy rate to 84% when aiming to identify native speakers of Arabic, English, and Spanish.

Ethnicity is the fact or state of belonging to a social group that has a *common national* or *cultural tradition* (Isajiw, 1993). Ethnicity can be fairly reliably predicted by inputting an individual's surname, geolocation, and other related attributes (e.g., gender and age) (Imai and Khanna, 2016). By using this approach, one can further determine his/her native language. A Bayesian method has been utilized to compute the posterior probability of each ethnic category for any given individual in this algorithm. Torvik and Agarwal (2016) is another typical work that developed a novel approach to identify a scholar's first language by inputting his/her full name. This algorithm, as well as their previous work (Smith, Singh, & Torvik, 2013) upon gender prediction, includes the whole PubMed author information and involves a nearest neighbor approach. The output of this algorithm includes a quantitative probability estimate of a given scholar's ethnicity (e.g., English, German). The current work adopts their proposed algorithm to identify the scholars' most probable ethnicities.

# 3. METHODOLOGY

The road map for this study is shown in Figure 1. First, the data set of this study is introduced. Then, author information is extracted for ethnicity classification and decision of the ethnicity of each manuscript. Next, full-text articles (XML format) are preprocessed with Python scripts. We start this step by extracting all text within the tag <p> from the full-text with *re* and *xml* and then remove the remaining tags and tokenized sentences with *NLTK* (Bird, Klein, & Loper, 2009) when abbreviations are replaced by their complete forms, i.e., "*et al.*" In order to calculate the linguistic features from two perspectives (syntactic and lexical complexity), Stanford Parser (Dan & Christopher, 2003) is applied to do POS tagging. Tregex[1] is used to extract clauses according to (Lu, 2010). When calculating measures of lexical complexity, we merge the POS tags given by Tree Bank. For instance, "NN" and "NNS", etc., are all counted as nouns. Finally, the manuscripts are grouped by the ethnicities of the authors and their linguistic features mapped for further analysis.

---

[1] https://nlp.stanford.edu/software/tregex.shtml

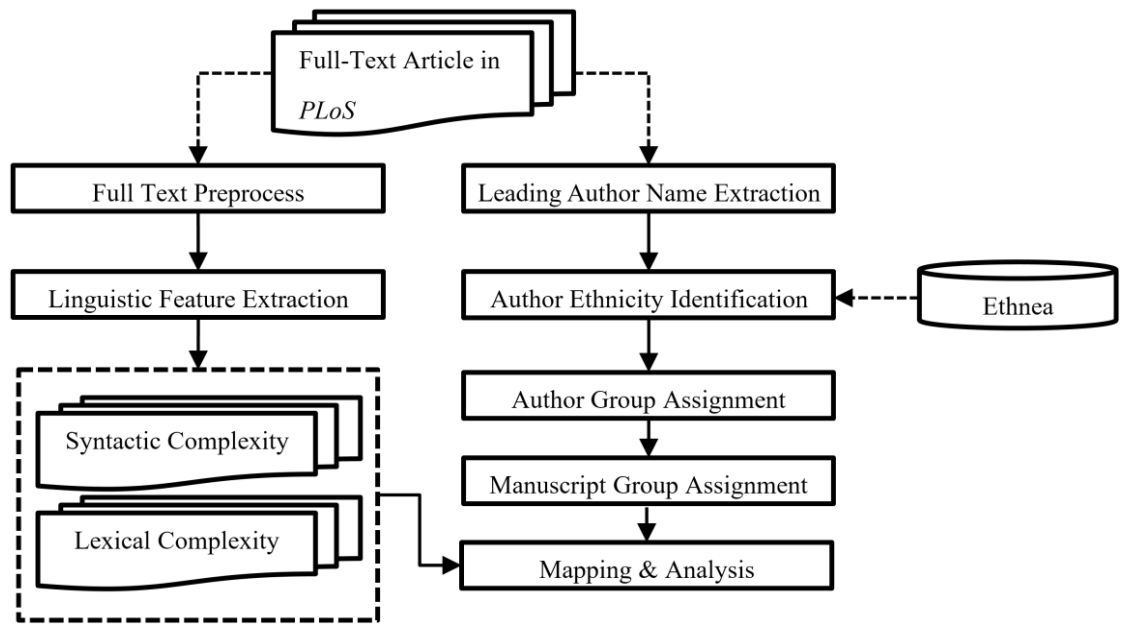

**Figure 1. Road map for this study.**

## *Data*

The data in this study consist of 172,662 full-text articles published from 2006 to 2015 in the *PLoS* journal family, a set of peer-reviewed journals covering various disciplines (detailed in Table 2). Of these, 9,282 articles pre-labeled by *PLoS* as non-research articles are removed from our data set.

## *Author-Level Ethnicity Identification*

There are various authorship practices across journals, disciplines, and fields. For example, the *PLoS* journal families all follow the authorship criteria proposed by International Committee of Medical Journal Editors (ICMJE) [2] and come up with a relatively sophisticated taxonomy of author contribution.

**Table 2. The distribution of article numbers among journals.**

| Journal Name | # of Publications | Ratio (%) |
|:---:|:---:|:---:|
| *PLoS Clinical Trials[3]* | 68 | 0.04 |
| *PLoS Medicine* | 2,966 | 1.72 |
| *PLoS Biology* | 4,023 | 2.33 |
| *PLoS Neglected Tropical Diseases* | 4,139 | 2.40 |
| *PLoS Computational Biology* | 4,334 | 2.51 |
| *PLoS Pathogens* | 5,123 | 2.97 |
| *PLoS Genetics* | 5,718 | 3.31 |
| *PLoS One* | 146,291 | 84.72 |
| **Total** | **172,662** | **100.00** |

According to the authorship policy of *PLoS* journals, writing contribution includes two aspects: drafting and revision. We randomly sampled 1,000 articles from the full data set and parsed the author contribution of each manuscript. We found that the first authors (FAs) and the corresponding authors (CAs) usually (more than 90%) play a role in these two aspects of writing contribution. Therefore, this study focuses upon FAs and CAs, which we will collectively call "leading authors" (LAs) as they greatly impact the writing style of manuscripts published as *PLoS* articles. Despite strong contributions to other parts of a study, the remaining authors influence the writing to a much lesser degree.

Thus, we extract the information of all the LAs from each manuscript to determine their ethnicities. Using these names, we employ Ethnea (Torvik & Agarwal, 2016) to identify the potential ethnicity of each LA. Ethnea is a system applied to predict ethnicity based on the geo-temporal distribution of names of authors in PubMed (articles in *PLoS* are indexed by PubMed), DBLP, MAG, ADS, NIH, NSF, and USPTO. In Ethnea, ethnicities are assigned to given names based on a large-scale and dense set of instances, which are geocoded and mapped by MapAffil (Torvik, 2015) using authors' affiliation information. Ethnea provides 26 kinds of ethnicities, covering nearly 99.7% authors in PubMed, including English, Hispanic, Chinese, German, Japanese, French, Italian, Slav, Indiana, Arab, Korean, Vietnamese, Nordic, Dutch, Turkish, Israeli, Greek, African, Hungarian, Thai, Romanian, Baltic, Indonesian, Caribbean, Mongolian, and Polynesian. Ethnea labels an author with only one ethnicity if the probability of the ethnicity > 60% and no other ethnicity is > 20%. Otherwise, if the total probability of the top two ethnicities is > 60% and no other ethnicity > 20%, Ethnea picks the top two. But if there are more than two ethnicities > 20%, then Ethnea labels the ethnicity of the author "UNKNOWN". According to Torvik & Agarwal (2016), Ethnea achieved a high level of agreement (78%) with EthnicSeer (Treeratpituk & Giles, 2012) on a dataset of 4.7 million authors in PubMed. Ye *et al.* (2017) reported that Ethnea obtained better performances (F1=0.58) than other systems, e.g., HMM (Ambekar *et al.*, 2009) (0.364), and EthnicSeer (0.571) on Wikipedia data and Email/Twitter Data.

---

[2] http://journals.plos.org/plosone/s/authorship

[3] *PLOS Clinical Trials* was merged in August 2007 with *PLoS One.*

4,569 articles without corresponding authors are removed from the dataset; 8,034 articles that contain authors with unknown ethnicities are also removed. Therefore, 150,777 articles comprise our final data set.

*Author Classification Strategy*

Using the authors' Ethnicity information, we assign each of them to one of the two groups (in Table 3): Group A (English ethnicity) and Group B (non-English ethnicity, e.g., German and Chinese).

**Table 3. Annotation schema.**

| Ethnic group | Label | Ethnicity |
|---|---|---|
| English | A | English |
| Other | B | French, Dutch, German, Hispanic, Italian, Turkish, Slav, Romanian, Greek, Baltic, Caribbean, Nordic, Indian, Japanese, Chinese, Arab, Israeli, Korean, African, Vietnamese, Hungarian, Thai, Indonesian, Mongolian |

*Manuscript Classification*

We group articles based on the classification of its LA(s) given that LAs make more contribution to writing and editing manuscripts. Table 4 demonstrates the strategy of group assignment with the distribution of each paper by group. The assignment implies that LAs equally contribute to writing.

**Table 4. Group assignment strategy for manuscripts.**

| Group of FA | Group of CA | Group of Article | # of article |
|---|---|---|---|
| A | A | A | 18,055 |
| A | B | AB | 32,318 |
| B | A | | |
| B | B | B | 100,404 |

*PLoS* is supposed to classify every publication with at least one subject area based on its taxonomy[4]. However, only 118,261 articles out of 150,777 (78.4%) are given discipline tags in the XML files, belonging to 8,131 unique subject areas covering both natural sciences and social sciences. The top 20 subject areas are provided in Table 5, making up 85.7% of the 118,261 publications and 67.3% of the final data set (150,777).

[4] http://journals.plos.org/plosone/s/help-ussing-this-site#loc-subject-areas

**Table 5. Top 20 subject areas in this study.**

| Subject Area | # of Publications | Rate |
|---|---|---|
| Biology | 61,658 | 0.52 |
| Medicine | 42,359 | 0.36 |
| Biochemistry | 23,025 | 0.19 |
| Molecular cell biology | 21,639 | 0.18 |
| Biology and life sciences | 21,338 | 0.18 |
| Genetics | 19,405 | 0.16 |
| Microbiology | 17,327 | 0.15 |
| Genetics and Genomics | 15,543 | 0.13 |
| Neuroscience | 14,957 | 0.13 |
| Medicine and health sciences | 14,396 | 0.12 |
| Computational biology | 14,209 | 0.12 |
| Infectious diseases | 13,492 | 0.11 |
| Model organisms | 13,397 | 0.11 |
| Immunology | 13,179 | 0.11 |
| Anatomy and physiology | 11,549 | 0.10 |
| Animal models | 10,962 | 0.09 |
| Physiology | 10,959 | 0.09 |
| Oncology | 10,756 | 0.09 |
| Epidemiology | 9,844 | 0.08 |
| Ecology | 9,824 | 0.08 |

*Measuring Scientific Writing Style Using Language Complexity*

Syntactic complexity focuses on the sentence-level complexity of language performance while the lexical complexity quantifies the level of vocabulary. Researchers have developed several indicators with specific quantitative variables to measure the two aspects of complexity, summarized in Table 6. Syntactic Complexity, as Ortega (2003) describes, "(also called syntactic maturity or linguistic complexity) refers to the range of forms that surface in language production and the degree of sophistication of such forms. This construct is important in second language research because of the assumption that language development entails, among other processes, the growth of an L2 (Second-language) Learner's syntactic repertoire and her or his ability to use that repertoire appropriately in a variety of situations" (p.492). Measurements for syntactic complexity can be divided into three sub-groups: sentence length, sentence complexity, and other measurements (e.g., number of sentence phrases) (Vajjala & Meurers, 2012). Lexical complexity, according to Laufer & Nation (1995), "attempt(s) to quantify the degree to which a writer is using a varied and large vocabulary. There has been interest in such measures for two reasons—they can be used to help distinguish some of the factors that affect the quality of a piece of writing, and they can be used to examine the relationship between vocabulary knowledge and vocabulary use" (p.307). Lexical complexity includes three sub-groups: lexical diversity, lexical sophistication, and lexical Density (Vajjala & Meurers, 2012).

**Table 6. Syntactic and Lexical complexity and their indicators.**

| Complexity | Indicators | Descriptions | Variables |
|---|---|---|---|
| Syntactic Complexity | Sentence Length | Number of words in a sentence unit (sentence, T-unit, or clause) | Sentence Length (e.g., Ferris, 1994b; Ortega, 2003) |
| | | | T-Unit Length (e.g., Vajjala & Meurers, 2012) |
| | | | Clause Length and its variations (e.g., Ferris, 1994a; Kormos, 2011) |
| | Sentence Complexity | Number of sentence phrases | Sentence Weight (DiStefano, & Howie, 1979) |
| | | | # of Clauses (Ferris, 1994a; Kormos, 2011) |
| | | | # of T-units Per Sentence (Ortega, 2003) |
| | | | # of Clauses Per Sentence (Ferris, 1994a) |
| | | | # of Clauses Per T-unit (Ortega, 2003) |
| | | | Depth of Modification (DiStefano, & Howie, 1979) |
| | Other | Number of sub-ordinations in the sentence | # of sentence phrases (e.g. # of NPs) (Vajjala & Meurers, 2012) |
| | | | Length of sentence phrases (e.g. length of NPs) (Vajjala & Meurers, 2012) |

| Lexical Complexity/ Richness | Lexical Diversity /Variation | Number of different words are used in the text | # of words and its variations (Ellis & Yuan, 2004) |
| | | | Type-Token Ratio (TTR) and its variations (e.g., Engber, 1995; Kormos, 2011) |
| | Lexical Sophistication | Degree of sophistication of lexical items | Word Length and its variations (e.g., Ferris, 1994b) |
| | | | Word List Coverage (Ellis & Yuan, 2004; Kormos, 2011) |
| | Lexical Density | Proportion of lexical items by the total number of tokens | PoS Ratio and its variations (Ellis & Yuan, 2004; Ojima, 2006) |

Using these existing indicators, this paper measures scholars' scientific writing from the perspective of Linguistic Complexity (Table 7), to achieve an understanding of the scholars' writing style and compare the difference between scholars of different ethnicities.

**Table 7. Indicators measuring scientific writing style in English.**

| Aspects | Indicators | Descriptions | Formulas |
|---|---|---|---|
| Syntactic Complexity | Sentence Length | Calculating average number of words in sentences of each article | $MSL = \frac{\sum_{i=1}^{N} SLi}{N}$ |
| | Sentence Complexity | Counting the number of clauses per sentences | $Clause\ Ratio = \frac{\#\ of\ all\ clauses}{\#\ of\ all\ sentences}$ |
| Lexical Complexity | Lexical Diversity | Type-Token Ratio in each article | $TTR = \frac{\#\ of\ Distinct\ words}{\#\ of\ tokens}$ |
| | Lexical Density | Counting the ratio of lexical items in tokens in each paper based on their part of speech (lexical class) | $Type\ Ratio = \frac{\#\ of\ Type\ items}{\#\ of\ Tokens}$ |
| | Lexical Sophistication | Counting the length of nouns, verbs, adjectives, and adverbs | $MWL = \frac{\sum_{i=1}^{N} WLi}{N}$ |

**<u>Syntactic Complexity</u>**. Sentence length and sentence complexity have been used as indicators to assess the syntactic complexity of NESs and NNESs because both indicators are beneficial for identifying improvement of NNESs or observing differences between NNESs and NESs (Kormos, 2011; Ojima, 2006). Sentence length is represented by the average number of words in a sentence of each paper (MSL in Table 7). Other similar variables (e.g., average T-unit/clause length) are not adopted because MSL is commonly used. The clause ratio of each article is calculated by dividing the number of clauses (i.e., a structure with a subject and a finite verb (Polio, 1997)) by the total number of sentences to measure sentence complexity, and this indicator has been used in various studies (e.g., Ferris, 1994a; Lu, 2010; Polio, 1997).

**Lexical Complexity**. Regarding lexical complexity, the indicators of lexical diversity, lexical density, and lexical sophistication are used to understand the difference in scientific writing between NESs and NNESs. Lexical diversity, assessed by the Type-Token Ratio (TTR) of each article (Engber, 1995), describes the total number of unique words normalized by the length of the text. Lexical sophistication, measured by the average length of words in each paper (MWL), reflects the cognitive complexity for both writers and readers (Juhasz, 2008). Word length can be calculated using two methods: number of characters or number of syllables in a word (Vajjala & Meurers, 2012). The former method is preferred for its frequent usage and reduced complexity of calculation. Lastly, lexical density is calculated by the ratio of lexical items (i.e., nouns, verbs, adjectives, and adverbs) to the total number of words (Lu, 2011). While lexical items provide semantic meaning, there are conflicting studies about whether using adjectives and adverbs (collectively called modifiers) improve or impair the readability of the text (Aziz, Fook, & Alsree, 2010; De Clercq & Hoste, 2016; DiStefano & Howie, 1979; DuBay, 2004; Vajjala & Meurers, 2012).

# 4. RESULTS

*Syntactic Complexity*

## <u>Sentence Length</u>

We plot the distribution of average lengths of sentences in the manuscripts for each group in Figure 2-A. Generally, the average sentence length of most manuscripts is longer than 25 words. Figure 2-A indicates that the more similar the ethnicity of manuscripts to English, the longer the sentences are. Average sentence length in Group A (28.2) is longer than that in Group B (26.9). Between the two groups lies Group AB (27.8), where manuscripts are collaborations by authors from Groups A and B. From the plot, we can also find that the differences between groups are marginal.

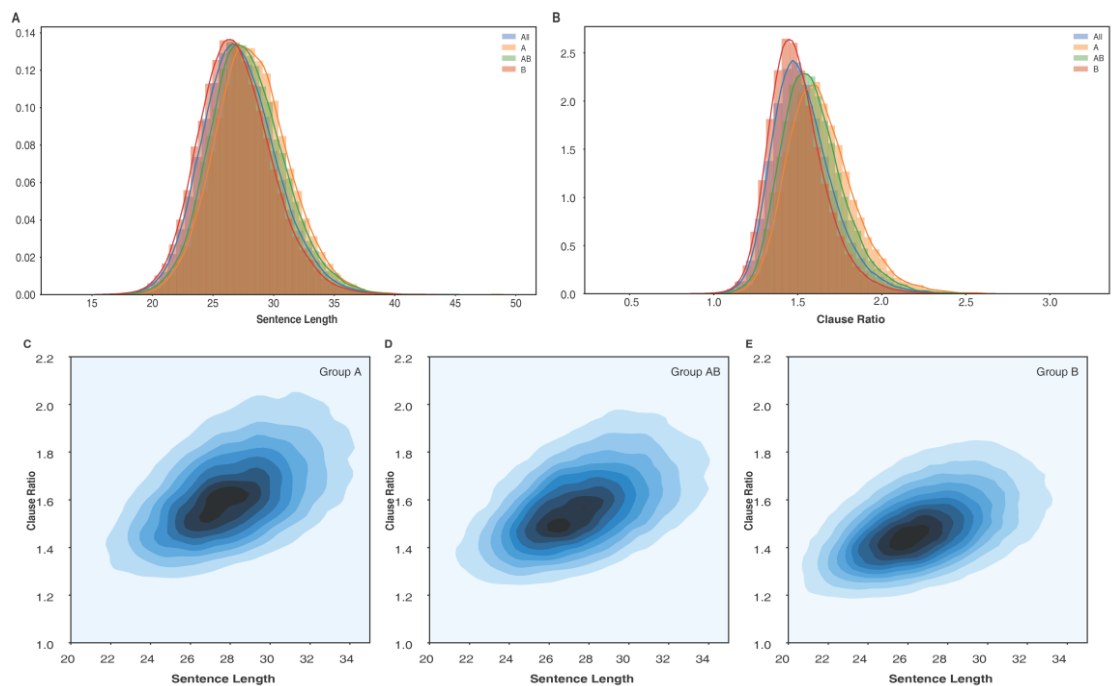

**Figure 2. Distribution of syntactic complexity features by group: A. Sentence length distribution; B. Clause ratio distribution; C. joint probability distribution plot of syntactic feature in Group A; D. joint probability distribution of syntactic feature Group AB; and E.**

**joint probability distribution of syntactic feature in Group B.**

## Sentence Complexity

Figure 2-B shows the distributions of clause ratios according to different groups. Articles in Group A tend to use more clauses (1.64 clauses per sentence), followed by Groups AB (1.59) and B (1.50), which show a similar trend in sentence length. For the cross-lingual group, the collaboration between LAs with different ethnic backgrounds also moderate the usage of clauses in scientific writing: ABs achieved a higher average clause ratio than those from Group B (1.5) but lower than those by A.

The joint plots for each group, shown in subplots C-E of Figure 2, also suggest that there is a positive correlation between average sentence length and clause ratio: on average, the longer the sentences are, the more clauses there are in the manuscripts. The disparity between groups suggests that given a certain average sentence length, manuscripts in Group A are more likely to employ more clauses in their sentences than those in Group AB, which are followed of Group B; and *vice versa*.

*Lexical Complexity*

## Lexical Diversity

We recognize that the length of articles is critical to the results of Type-Token Ratio (e.g., Tweedie & Baayen, 1998); so TTR values for each group are compared according to the length range of manuscripts with 6,000 to 10,000 words shown in Figure 3. On average, TTRs of all groups are greater than 20%. However, between groups, there is no clear pattern observed that manuscripts from three groups show no differences in employing diverse vocabulary.

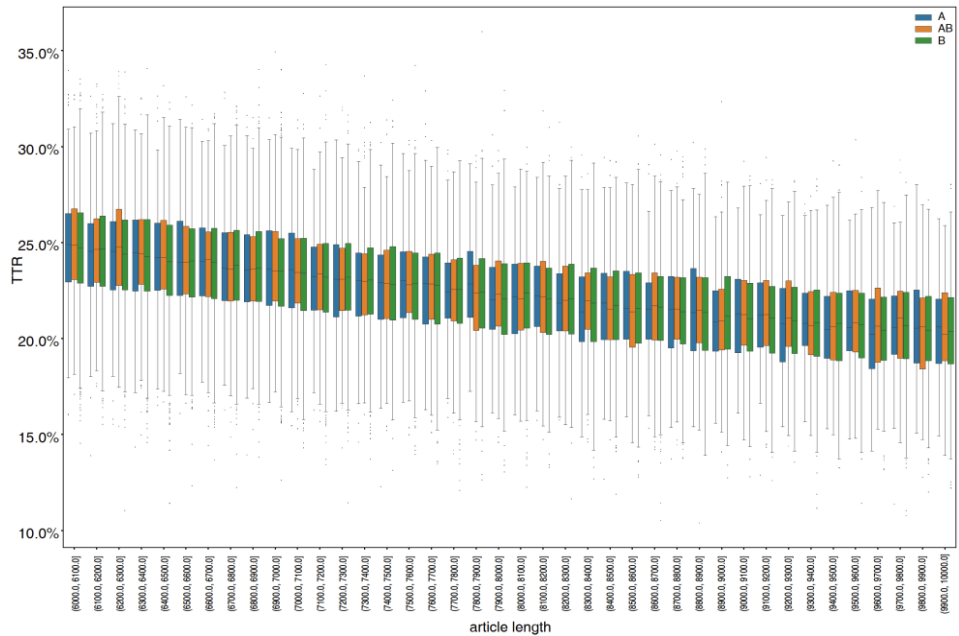

**Figure 3. TTR distribution by article length (partial, see the complete results in the Appendix Figure A1).**

## Lexical Sophistication

Figure 4 shows the distributions of average lexical word length by group respectively. Generally, average length of nouns (6.66) is longer than that of verbs (6.25) and shorter than that of adverbs (6.77). Average length of adjectives is the longest of all (7.61).

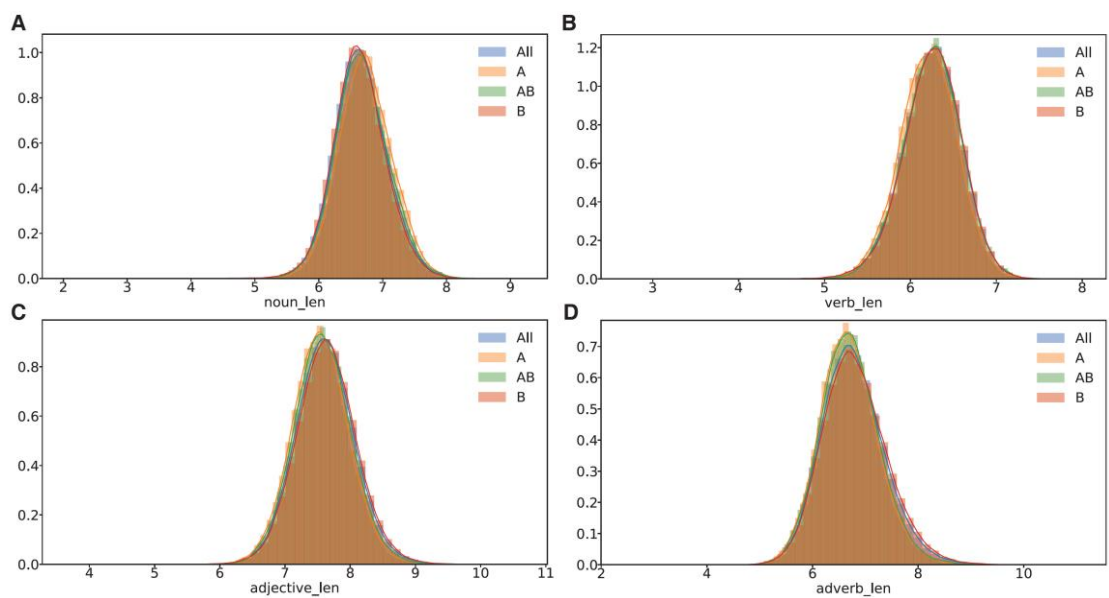

**Figure 4. Distributions of lexical sophistication by group: A. average noun length; B. average verb length; C. average adjective length; and D. average adverb length.**

Specifically, on average, manuscripts in Group A continue to use marginally longer nouns than those in Group AB (6.73 vs 6.68), which is followed by those in Group B (6.63). Manuscripts in Group AB show a moderate average noun length between the monolingual groups (A and B). By contrast, the patterns of verbs are the opposite: manuscripts Group A use marginally shorter verbs (6.21) than the other two groups, which show almost equivalent average verb lengths (6.25), which echoes findings by Ferris (1994a). Indicated in Figure 4-C, Group A uses the shortest adjectives (7.54), followed by Groups AB (7.58) and B (7.63), in which average adjective length can be marginally adjusted by LAs from different ethnic backgrounds. Similarly, average length of adverbs shows the same pattern with that of adjectives: Group C also uses the longest adverbs (6.68), followed by Groups AB (6.71) and B (6.81).

## Lexical Density

On average, we find that manuscripts are made up of 37% nouns, 15% verbs, 10%

adjectives, and 3% adverbs (lexical density is only measured by lexical items, while other types of words, e.g., preposition, are not considered in this study); among the lexical items, nouns are used most frequently. Figure 5 shows the distribution of the usage of the lexical items by group respectively.

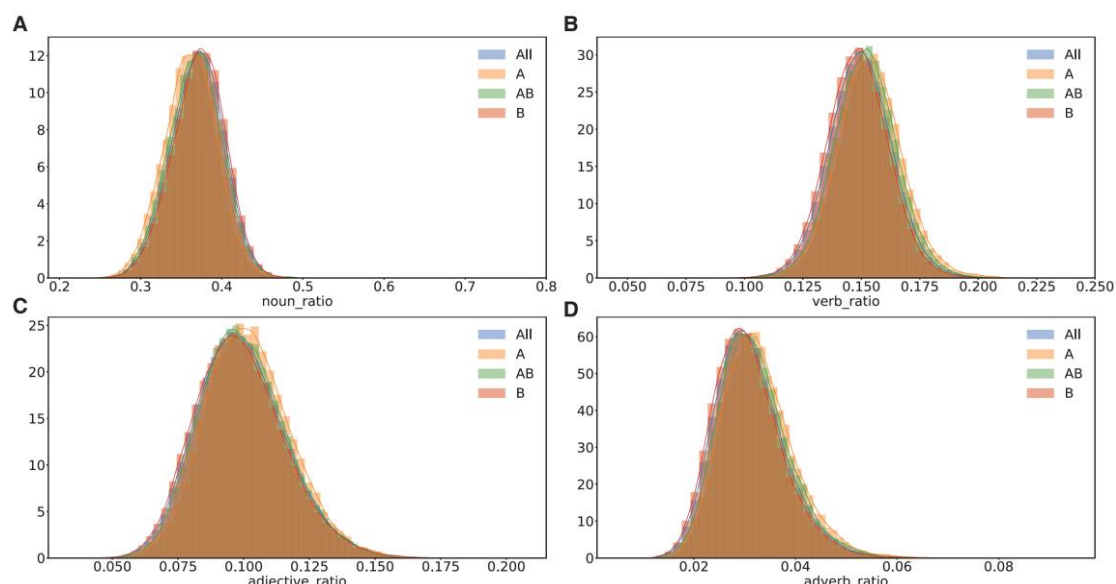

**Figure 5. Distributions of lexical density by group: A. noun ratio; B. verb ratio; C. adjective ratio; D. adverb ratio.**

Figure 5-A displays the noun usage by group. Nouns make up 37.2% of manuscripts in Group B, the largest ratio among the three groups; while in Group A nouns are 36.1%, the smallest ratio, in between which lies Group AB, 36.7%, collaborations between NNESs and NESs. In contrast, verb usage (in Figure 5-B) shows a reverse trend: Group A uses the verbs most frequently (15.4%), which is followed by Groups AB (15.2%) and B (14.8%) accordingly. Figure 5-C shows that Group A employs adjectives most frequently (10.2% on average), closely followed by Groups AB and B in order (10.0% respectively). The differences between groups are even smaller. Similarly, Figure 5-D suggests that the three groups obtain similar average adverb ratio (around 3%).

Similarly to lexical sophistication, manuscripts from the three groups show marginal differences in lexical density especially in adjective and adverb usages.

## 5. DISCUSSION

*Marginal Differences in Linguistic Complexity of Scientific Writing*

Considering syntactic complexity, authors from Group A employ longer and more complex sentences than those in Group B. Our result indicates that the average sentence lengths are quite close between groups, and the margins between Groups A and AB or between AB and B are even narrower. Interestingly, authors from Group A are capable of writing both longer and shorter sentences while those from other groups show less flexibility in writing sentences in varying lengths.

Regarding lexical sophistication, most lexical items comprise six to seven characters. Adverbs are slightly longer (6.5-7.5 characters), which is common in English. These short lexical items, according to related studies (Juhasz, 2008), reduce the cognitive pressure on readers and thus improve the readability of manuscripts.

Scholars in Group A generally use longer nouns, shorter verbs, and shorter modifiers in contrast to those in Group B. The LAs from Group A apply longer nouns in their writing probably because of their larger vocabulary, especially for native English scholars. One possible reason is that frequent adoption of nominalizations is a conventional way to show scholars' politeness and objective position (Billig, 2008); and the adoption also indicates the sophistication of NNESs' language skills suggested by Wang & Chen (2008). The authors of Group A usually use shorter verbs, while those in Group B usually employ longer verbs. One possible reason is that their ethnic similarity to NEWs enables those in Group A to find short or simple verbs from a larger vocabulary (or simple verb phrases) to convey their ideas precisely, while authors in Group B may use more verbalizations to express their ideas with limited vocabularies. Frequent use of nominalizations leading to shorter verb phrases (e.g., "give approval of…" vs. "approve…") may be another possible reason for this difference.

On lexical density, the results indicate that authors from Group A use slightly fewer nouns and more verbs than those in Group B. No differences have been found in using modifiers since these kinds of words are not as frequently adopted in scientific writing.

However, from the observations we have found that the differences between groups are marginal across the features of linguistic complexity, especially the lexical complexity. This is in accordance with observations of other studies in linguistics (i.e., Dobao, 2012; Ellis & Yuan, 2004; Engber, 1995; Ferris, 1994a, b; Kormos, 2011; Ortega, 2003), comparing writings between NEWs and NNEWs within different scenarios and study settings, that the differences between more proficient writers and less proficient ones can be "marginally statistically significant" (Engber, 1995). In other words, the differences might not be as practically significant as suggested by Ortega (2003).

Given the *de facto* barriers for NNESs posed by the difficulties of English scientific writing, the marginal differences of linguistic complexity in scientific writing may not be sufficient to fully capture the language barrier of NNESs.

*Cross-lingual Collaboration Might Benefit the NNESs*

Despite of the marginal difference, these findings might direct us to think further about cross-language collaboration. For example, for the sentence length, articles from Group AB (collaborations by LAs from different groups) ranked in the middle, compared to manuscripts in Groups A and B. It implies that authors from Group A can help those in Group B write longer sentences. Similar observations can be found in other features: the manuscripts of Group AB adopt a higher clause ratio, longer nouns, shorter verbs, and shorter modifiers than those in Group B. This suggests that cross-lingual collaboration can help NNESs improve their scientific writing. This fits with past and present studies which discuss ways for NNESs to collaborate with NESs so that both parties can benefit (Lee & Bozeman, 2005; Lee, Sugimoto, Zhang, & Cronin, 2013).

Organizations also provide ways to facilitate cross-language collaboration. For example, since 2016, the Annual Meeting of the Association for Information Science and Technology (ASIS&T) has provided detailed revision suggestions for early submissions and opened up professional and academic mentoring program to help scholars withstudy design, academic career planning, etc.

*Limitations*

This study only considers the contribution of the FAs and CAs as LAs in English scientific writing and ignores the possible contribution of other authors. This limits our understanding of the effects from other non-leading collaborators in scientific writing. One possible solution is to mine the specific contributions of the collaborators addressed by the authors mentioned in the end of the manuscript in order to achieve a deeper understanding.

The corpora of this study are mainly taken from hard sciences. Future investigations may require insight into other domains—especially social sciences—to achieve a comprehensive understanding of scientific writing style. A possible solution to investigate subject-specific writing style in our data set is also of value given that different topics tend to demonstrate diverse linguistic patterns (Yang, Lu, & Weigle, 2015). However, the papers published in *PLoS* are usually multidisciplinary studies, and thus attempting to classify them by their subjects could introduce bias.

This paper takes the authors whose names are labeled as "English" (ethnicity) by Ethnea as the most similar to native English speakers, which might reduce the differences between the real-world NESs and NNESs. Simultaneously, the precision of Ethnea can be another factor that affects the result of this study. The proper way to fully overcome the limitation is to survey the authors about their linguistic proficiency and their native languages—for instance, we can divide authors into more groups to compare the differences and to study the effectiveness of collaboration on scientific writing.

## 6. CONCLUSION

This study investigates the differences between authors from different ethnic backgrounds in the linguistic complexity of scientific writing. Via the 12 linguistic features of more than 150,000 full-text articles, we have found that the authors with the English ethnic background usually produce longer sentences comprising more clauses, employ longer nouns and shorter verbs, and utilize more verbs and fewer nouns. However, the differences are marginal. Taking into account the difficulty the NNESs are facing in English scientific writing, we come up with several ways that might be helpful for them to improve their writing in the long run, especially via collaboration with NESs.

There are many related studies that can be implemented in the future. First, this paper mainly focuses on a broad view on differences between groups of scholars clustered by ethnic similarities; a more nuanced investigation should be conducted on linguistic differences between sections within each specific paper, which would enable us to examine how authors show different writing styles in the same paper. Second, according to the *PLoS* publisher's publishing policy, every paper before publication should be clear and error-free; and if necessary, authors can refer to professional editing service for help. Despite that the professional editing is not a must, authors, especially NNES, may be prone to ask for help from companies or individuals, which might bring survivorship bias to our findings. Therefore, purely quantitative investigation of the data might be insufficient. Future researchers can therefore involve more qualitative strategies, such as questionnaire and interview, to investigate authors in our data set and explore how they improve their papers before submission. Meanwhile, conference papers could be a promising data source to investigate in the future, since the editorial processing for conference papers is not so restricted as that for journal papers. Furthermore, we will also dive deep into the semantic level of scientific writing to understand the writing style of NESs and NNESs and hopefully find improved ways to help NNESs in scientific writing.

## ACKNOWLEDGEMENTS

This work is supported in part by Major Projects of The National Social Science Fund of China (No. 16ZAD224) and China Scholarship Council (ID: 201606840093). The authors would like to thank Markus Dickinson and Gregory Maus for their precious suggestions on this study and to appreciate the IT center in School of Economics and Management, Nanjing University of Science and Technology and Xu Jin for technical supports. The authors are also grateful to all the anonymous reviewers for their precious comments and suggestions.

# REFERENCES

Ambekar, A., Ward, C., Mohammed, J., Male, S., & Skiena, S. (2009). Name-ethnicity classification from open sources. In *Proceedings of the 15th ACM SIGKDD international conference on Knowledge Discovery and Data Mining* (pp. 49-58), June 28-July 1, 2009, Paris, France.

Aziz, A., Fook, C. Y., & Alsree, Z. (2010). Computational text analysis: A more comprehensive approach to determine readability of reading materials. *Advances in Language and Literary Studies, 1*(2), 200-219.

Bailey, M. (2011). Science editing and its effect on manuscript acceptance time. *American Medical Writers Association Journal, 26*(4), 147-152.

Beers, S.F., & Nagy, W.E. (2009). Syntactic complexity as a predictor of adolescent writing quality: Which measures? Which genre?. *Reading and Writing*, *22*(2), 185-200.

Bergsma, S., Post, M., & Yarowsky, D. (2012). Stylometric analysis of scientific articles. In *Proceedings of the 2012 Conference of the North American Chapter of the Association for Computational Linguistics: Human Language Technologies* (pp. 327-337), July 8-14, 2012, Jelu Island, South Korea.

Billig, M. (2008). The language of critical discourse analysis: The case of nominalization. *Discourse and Society, 19*(6), 783-800.

Bird, S., Klein, E., & Loper, E. (2009). *Natural language processing with Python: analyzing text with the natural language toolkit*. Newton, MA: O'Reilly Media, Inc.

Brants, T. (2000). TnT: A statistical part-of-speech tagger. In *Proceedings of the Sixth Conference on Applied Natural Language Processing* (pp. 224-231), April 29-May 4, 2000, Seattle, Washington, U.S.A.

Brown, P.F., Pietra, V.J.D., Pietra, S.A.D., & Mercer, R.L. (1993). The mathematics of statistical machine translation: Parameter estimation. *Computational Linguistics, 19*(2), 263-311.

Campbell, D.A., & Johnson, S.B. (2001). Comparing syntactic complexity in medical and non-medical corpora. In *Proceedings of the AMIA Symposium* (pp. 90-94). November 3-7, 2001, Washington D.C., U.S.A.

Chandler, J. (2003). The efficacy of various kinds of error feedback for improvement in the accuracy and fluency of L2 student writing. *Journal of Second Language Writing, 12*(3), 267-296.

Crowhurst, M. (1991). Interrelationships between reading and writing persuasive discourse. *Research in the Teaching of English, 25*(3), 314-338.

Curry, M.J., & Lillis, T. (2004). Multilingual scholars and the imperative to publish in English: Negotiating interests, demands, and rewards. *TESOL Quarterly, 38*(4), 663-688.

Dan, K. & Christopher, D. (2003). Accurate unlexicalized parsing. In *Proceedings of the 41st Meeting of the Association for Computational Linguistics* (pp. 423-430), July 7-12, 2003, Sapporo, Japan.

De Clercq, O., & Hoste, V. (2016). All mixed up? finding the optimal feature set for general readability prediction and its application to English and Dutch. *Computational Linguistics, 42*(3), 457-490.

DiStefano, P., & Howie, S. (1979). Sentence weights: An alternative to the T-Unit. *English Education, 11*(2), 98-101.

Dobao, A.F. (2012). Collaborative writing tasks in the L2 classroom: Comparing group, pair, and individual work. *Journal of Second Language Writing, 21*(1), 40-58.

DuBay, W.H. (2004) The principles of readability. Retrieved 27 November 2017 from http://www.impact-information.com/impactinfo/readability02.pdf

Ellis, R., & Yuan, F. (2004). The effects of planning on fluency, complexity, and accuracy in second language narrative writing. *Studies in Second Language Acquisition, 26*(01), 59-84.

Engber, C.A. (1995). The relationship of lexical proficiency to the quality of ESL compositions. *Journal of Second Language Writing, 4*(2), 139-155.

Estival, D., Gaustad, T., Pham, S.B., Radford, W., & Hutchinson, B. (2007). Author profiling for English emails. In *Proceedings of the 10th Conference of the Pacific Association for Computational Linguistics* (pp. 263-272), September 19-21, 2007, Melbourne, Victoria, Australia.

Ferris, D.R. (2004). The "grammar correction" debate in L2 writing: Where are we, and where do we go from here? (and what do we do in the meantime…?). *Journal of Second Language Writing, 13*(1), 49-62.

Ferris, D.R. (1994a). Rhetorical strategies in student persuasive writing: Differences between native and non-native English speakers. *Research in the Teaching of English, 28*(1), 45-65.

Ferris, D.R. (1994b). Lexical and syntactic features of ESL writing by students at different levels of L2 proficiency. *TESOL Quarterly, 28*(2), 414-420.

Flowerdew, J. (1999). Problems in writing for scholarly publication in English: The case of Hong Kong. *Journal of Second Language Writing, 8*(3), 243-264.

Gebhardt, R. (1980). Teamwork and feedback: Broadening the base of collaborative writing. *College English, 42*(1), 69-74.

Holmes, D.I. (1994). Authorship attribution. *Computers and the Humanities, 28*(2), 87-106.

Huang, J.C. (2010). Publishing and learning writing for publication in English: Perspectives of NNES PhD students in science. *Journal of English for Academic Purposes, 9*(1), 33-44.

Imai, K., & Khanna, K. (2016). Improving ecological inference by predicting individual ethnicity from voter registration records. *Political Analysis, 24*(2), 263-272.

Isajiw, W. W. (1993). Definition and Dimensions of Ethnicity: A Theoretical Framework. In *Proceedings of the Joint Canada-United States Conference on the Measurement of Ethnicity*, April 1-3. Washington, D.C: U.S. Government Printing Office, 407-427.

Jang, S. (2007). A study of students' construction of science knowledge: Talk and writing in a collaborative group. *Educational Research, 49*(1), 65-81.

Jarvis, S., & Paquot, M. (2015). Native language identification. In S. Granger, G. Gilquin, & F. Meunier Ed., *Cambridge Handbook of Learner Corpus Research*, Cambridge University Press: Cambridge, 2015.

Juhasz, B.J. (2008). The processing of compound words in English: Effects of word length on eye movements during reading. *Language and Cognitive Processes, 23*(7-8), 1057-1088.

Kessler, G., Bikowski, D., & Boggs, J. (2012). Collaborative writing among second language learners in academic web-based projects. *Language Learning and Technology, 16*(1), 91-109.

Keys, C.W. (1994). The development of scientific reasoning skills in conjunction with collaborative writing assignments: An interpretive study of six ninth-grade students. *Journal of Research in Science Teaching, 31*(9), 1003-1022.

Koppel, M., Schler, J., & Zigdon, K. (2005, May). Automatically determining an anonymous author's native language. In *International Conference on Intelligence and Security Informatics* (pp. 209-217). Springer, Berlin, Heidelberg.

Kormos, J. (2011). Task complexity and linguistic and discourse features of narrative writing performance. *Journal of Second Language Writing, 20*(2), 148-161.

Laufer, B., & Nation, P. (1995). Vocabulary size and use: Lexical richness in L2 written production. *Applied Linguistics, 16*(3), 307-322.

Lee, C.J., Sugimoto, C. R., Zhang, G., & Cronin, B. (2013). Bias in peer review. *Journal of the American Society for Information Science and Technology, 64*(1), 2-17.

Lee, S., & Bozeman, B. (2005). The impact of research collaboration on scientific productivity. *Social Studies of Science, 35*(5), 673-702.

Lowry, P.B., Curtis, A., & Lowry, M.R. (2004). Building a taxonomy and nomenclature of collaborative writing to improve interdisciplinary research and practice. *Journal of Business Communication, 41*(1), 66-99.

Lu, X. (2010). Automatic analysis of syntactic complexity in second language writing. *International Journal of Corpus Linguistics, 15*(4), 474-496.

Lu, X. (2011). A corpus-based evaluation of syntactic complexity measures as indices of college-level ESL writers' language development. *TESOL Quarterly, 45*(1), 36-62.

Mason, L. (1998). Sharing cognition to construct scientific knowledge in school context: The role of oral and written discourse. *Instructional Science, 26*(5), 359-389.

Mayfield, T.L., & Jones, R. (2001). You're not from round here, are you? Naive Bayes detection of non-native utterance text. In *Proceedings of the Second Meeting of the North American Chapter of the Association for Computational Linguistics*, June 2-7, 2001, Pittsburgh, PA, U.S.A.

Mišak, A., Marušić, M., & Marušić, A. (2005). Manuscript editing as a way of teaching scientific writing: Experience from a small scientific journal. *Journal of Second Language Writing, 14*(2), 122-131.

Nisioi, S. (2015, April). Feature analysis for native language identification. In *International Conference on Intelligent Text Processing and Computational Linguistics* (pp. 644-657). Springer, Cham.

Ojima, M. (2006). Concept mapping as pre-task planning: A case study of three Japanese ESL writers. *System, 34*(4), 566-585.

Ortega, L. (2003). Syntactic complexity measures and their relationship to L2 proficiency: A research synthesis of college-level L2 writing. *Applied linguistics, 24*(4), 492-518.

Polio, C.G. (1997). Measures of linguistic accuracy in second language writing research. *Language Learning, 47*(1), 101-143.

Rabinovich, E., Nisioi, S., Ordan, N., & Wintner, S. (2016). On the similarities between native, non-native and translated texts. *arXiv preprint*, arXiv:1609.03204.

Scontras, G., Badecker, W., & Fedorenko, E. (2017). Syntactic complexity effects in sentence production. *Cognitive Science, 39*(3), 559-583.

Skehan, P. (2009). Modelling second language performance: Integrating complexity, accuracy, fluency, and lexis. *Applied Linguistics, 30*(4), 510-532.

Silva, T. (1993). Toward an understanding of the distinct nature of L2 writing: The ESL research and its implications. *TESOL Quarterly, 27*(4), 657-677.

Shaw, P. (1991). Science research students' composing processes. *English for Specific Purposes, 10*(3), 189-206.

Smith, B.N., Singh, M., & Torvik, V.I. (2013). A search engine approach to estimating temporal changes in gender orientation of first names. In *Proceedings of the 13th ACM/IEEE-CS joint conference on Digital libraries* (pp. 199-208), July 22-26, 2013, Indianapolis, IN, U.S.A.

Tomkins, A., Zhang, M., & Heavlin, W.D. (2017). Reviewer bias in single-versus double-blind peer review. *Proceedings of the National Academy of Sciences of the United States of America, 114*(48), 12708-12713.

Torvik, V.I. (2015). MapAffil: A bibliographic tool for mapping author affiliation strings to cities and their geocodes worldwide. *D-Lib Magazine : The Magazine of the Digital Library Forum, 21*(11-12), 10.1045/november2015–torvik.

Torvik, V.I., Agarwal, S. (2016) Ethnea: An instance-based ethnicity classifier based on geo-coded author names in a large-scale bibliographic database. Paper presented at *International Symposium on Science of Science,* March 22-23, 2016, Library of Congress, Washington D.C., U.S.A.

Treeratpituk, P. & Giles, C.L. (2012). Name-ethnicity classification and ethnicity-sensitive name matching. In *Proceedings of the 26th AAAI Conference on Artificial Intelligence* (pp. 1141–1147), July 22-26, 2012, Toronto, Ontario, Canada.

Tsvetkov, Y., Twitto, N., Schneider, N., Ordan, N., Faruqui, M., Chahuneau, V., ... & Dyer, C. (2013). Identifying the L1 of non-native writers: the CMU-Haifa system. In *Proceedings of the Eighth Workshop on Innovative Use of NLP for Building Educational Applications* (pp. 279-287).

Tweedie, F.J., & Baayen, R.H. (1998). How variable may a constant be? Measures of lexical richness in perspective. *Computers and the Humanities, 32*(5), 323-352.

Vajjala, S., & Meurers, D. (2012). On improving the accuracy of readability classification using insights from second language acquisition. In *Proceedings of the Seventh Workshop on Building Educational Applications Using NLP* (pp. 163-173), July 8-14, 2012, Jelu Island, South Korea.

Wang, L., & Chen, Q. (2008). A Corpus-based Study on Nominalizations in Argumentative Essays of Chinese EFL Learners. *Foreign Languages in China, 5*(5), 54-110.

Wong, S.M.J., & Dras, M. (2011). Exploiting parse structures for native language identification. In *Proceedings of the Conference on Empirical Methods in Natural Language Processing* (pp. 1600-1610), July 27-31, 2011, Edinburgh, U.K.

Xu, J., Ding, Y., Song, M., & Chambers, T. (2016). Author credit-assignment schemas: A comparison and analysis. *Journal of the Association for Information Science and Technology, 67*(8), 1973-1989.

Yang, W., Lu, X., & Weigle, S. C. (2015). Different topics, different discourse: Relationships among writing topic, measures of syntactic complexity, and judgments of writing quality. *Journal of Second Language Writing*, *28*, 53-67.

Yang, Y.H., Marslenwilson, W.D., & Bozic, M. (2017). Syntactic complexity and frequency in the neurocognitive language system. *Journal of Cognitive Neuroscience, 29*(9), 1605-1620.

Yarrow, F., & Topping, K.J. (2001). Collaborative writing: The effects of metacognitive prompting and structured peer interaction. *British Journal of Educational Psychology, 71*(2), 261-282.

Ye, J., Han, S., Hu, Y., Coskun, B., Liu, M., Qin, H., & Skiena, S. (2017, November). Nationality classification using name embeddings. In *Proceedings of the 2017 ACM on Conference on Information and Knowledge Management* (pp. 1897-1906), November 06-10, 2017, Singapore, Singapore.

Zhang, C., Bu, Y., Ding, Y., & Xu, J. (2018). Understanding scientific collaboration: Homophily, transitivity, and preferential attachment. *Journal of the Association for Information Science and Technology, 69*(1), 72-86.

# APPENDIX

We provide a Kolmogorov-Smirnov test for all linguistic complexity features between different groups, as shown in Table A1. The values in this table are *p*-values corresponding to the test result of the certain features on any given two groups, indicating that the differences of the features among groups are statistically significant or not. We also provide the TTR distribution by article length in Figure A1.

**Table A1. *p* values of Kolmogorov-Smirnov Tests of linguistic complexity features between groups.**

| Features | Groups | | |
|---|---|---|---|
| | (A, B) | (A, AB) | (AB, B) |
| sentence length | 6.1E-155 | 1.3E-18 | 9.82E-74 |
| clause ratio | 0 | 5.66E-51 | 8E-206 |
| TTR | 6.21E-21 | 0.092551 | 3.7E-30 |
| noun length | 8.45E-30 | 6.62E-16 | 6.25E-08 |
| verb length | 9.84E-15 | 3.89E-15 | 0.920293 |
| adjective length | 1.03E-37 | 1.67E-07 | 1.47E-13 |
| adverb length | 4.99E-31 | 0.005609 | 1.02E-21 |
| noun ratio | 7.92E-94 | 1.74E-27 | 4.44E-23 |
| verb ratio | 6.2E-120 | 2.28E-13 | 5.79E-59 |
| adjective ratio | 5.44E-37 | 5.53E-12 | 6.27E-10 |
| adverb ratio | 2.05E-74 | 1.32E-09 | 1.18E-32 |

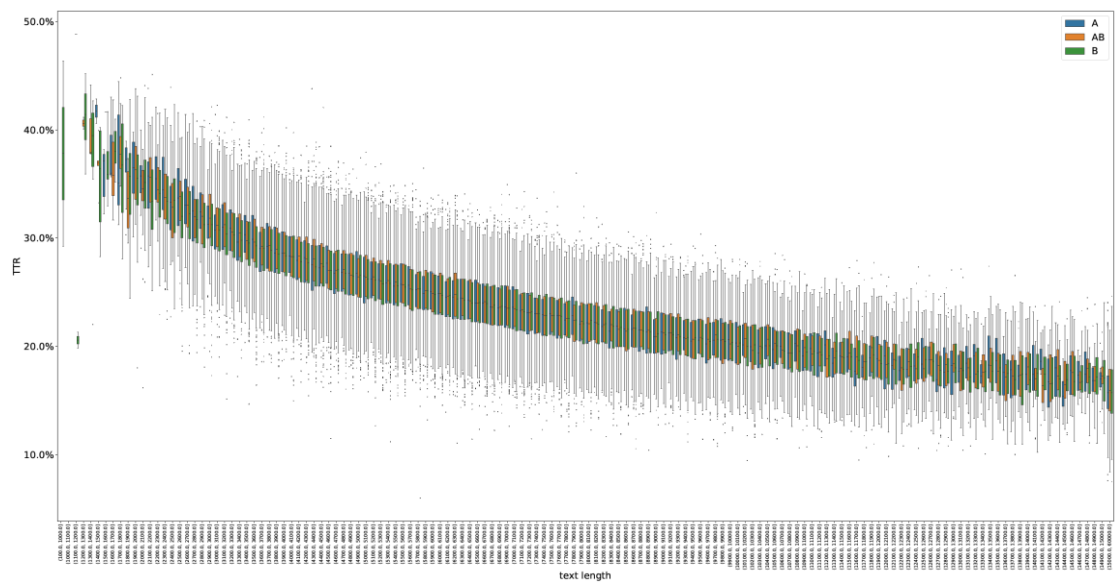

**Figure A1. TTR distribution by article length.**